\documentclass[11pt]{article}

\usepackage[preprint]{acl}
\usepackage{pifont}
\usepackage{times}
\usepackage{latexsym}
\usepackage[T1]{fontenc}
\usepackage[utf8]{inputenc}
\usepackage{microtype}
\usepackage{inconsolata}
\usepackage{graphicx}

\usepackage{amsmath}
\usepackage{amssymb}
\usepackage{booktabs}
\usepackage{algorithm}
\usepackage{algorithmic}

\title{ArcRAG: Agentic Compositional Retrieval over Knowledge Graphs}

\author{
  Jyotin Goel\thanks{Corresponding author.} \quad Abhinav Arun \quad Tejas Prakash Agarwal \\
  \textbf{Bhaskarjit Sarmah} \quad \textbf{Stefano Pasquali} \\
  Domyn \\
  \texttt{jyotin.goel@domyn.com}
}
\begin{document}

\maketitle

\begin{abstract}
Graph-based retrieval-augmented generation (GraphRAG) answers questions by
retrieving over a knowledge graph built from a corpus. Yet retrieval
strategies are designed around the structure their own construction pipeline
produces, and whether they survive a change of construction is rarely tested.
We study this cross-construction portability: whether one retrieval
procedure, with no construction-specific schema, relation names, or retuning,
stays effective when the graph is built differently. We introduce ArcRAG,
which induces a query-local graph interface at inference time (the node
labels, relation descriptors, and property keys reachable from the entities a
question names), plans multi-hop traversals against it, and executes and
repairs them as Cypher. We build each of the MuSiQue, HotpotQA, and TriviaQA
corpora as a schema-guided graph (typed nodes, a closed relation set, source
provenance) and as an open-information-extraction graph (untyped nodes,
free-text relations), with the retriever's configuration fixed across them.
We compare against a construction-agnostic Direct-Cypher agent, ablate each
stage across retry and pass budgets on both graph families, and compare
with results reported for prior GraphRAG systems, most of which design
retrieval together with their own graph construction. Across the two
constructions, we find that retrieval need not be co-designed with the graph:
query-time interface induction keeps a construction-agnostic retriever
effective when the construction changes, although construction still shapes
absolute accuracy.
\end{abstract}

\section{Introduction}
\label{sec:intro}

Retrieval-augmented generation (RAG) grounds a language model in an external
corpus by retrieving information pieces (passages) most similar to a query and conditioning
generation on them. Similarity ranking is sufficient when the answer is
contained in a single passage, and insufficient when the answer must be
composed across several. Consider a question of the form \emph{which member
of a given film's cast has two children who attended a particular
university}: answering it requires enumerating a set, testing a property of
each of its elements, and verifying a constraint that no individual passage
asserts. Top-$k$ alone cannot perform these operations. Graph-based RAG
addresses this by retrieving over a knowledge graph (KG) rather than a flat
collection of passages, so that the relations connecting evidence are part of
what retrieval returns. Composition, though, is only half of what these
settings demand: in regulated domains --- finance, law, healthcare,
compliance --- questions are seldom one-hop, and an answer must also be
\emph{traceable} to the specific disclosure, clause, or record it rests on. Success in these
settings is evidence returned together with the path that connects it,
rather than the $k$ passages nearest a query.

Dense embedding retrieval, the workhorse of naive RAG, cannot supply that,
and the reason is structural rather than a matter of scale. Each passage is
represented independently, while a multi-hop question depends on the
relations \emph{among} passages. As corpora grow and reasoning chains
lengthen, the relevant facts remain individually retrievable yet become
\emph{relationally invisible}: a vector index surfaces the constituent
pieces without representing the path that joins them, leaving the generator
to reconstruct those connections from semantically similar text amid
substantial surrounding noise. FinReflectKG--MultiHop demonstrates that this
is not a deficiency stronger models will absorb: even reasoning-oriented
models perform better given precisely KG-linked evidence than given
text-only page windows that contain the same facts in less structured
context \citep{arun2025finreflectkgmultihop}. Knowledge graphs close this
representational gap directly. Typed edges turn ``which entity connects to
which'' into a traversable structure, and the traversal followed during
retrieval can itself serve as provenance. This is the insight behind
GraphRAG \citep{edge2024graphrag} and the class of graph-based retrieval
systems it inspired, and it reduces the open problem to a single question:
given a knowledge graph, \emph{how should one retrieve over it?}

The difficulty in answering that question is that a knowledge graph is not
one kind of object. Graphs differ in \emph{schema commitment}, from open
information extraction with free-text relation strings to schema-guided
extraction against a closed, typed relation set, and in whether edges carry
\emph{provenance} back to a source chunk. Most GraphRAG systems fix one point
in this space and design retrieval for it, and cross-construction
portability is rarely tested directly.

Instead, each fixes a point in that space and builds a strategy that
presupposes it. Personalized PageRank presupposes a dense synonymy-linked
entity graph, and returns a ranked passage list rather than a traversal a user
can cite \citep{hipporag}; GraphRAG's global search presupposes a hierarchical
Leiden partition and the community summaries precomputed over it
\citep{edge2024graphrag}; LightRAG's dual-level retrieval presupposes an
entity-and-relation keyword index and remains a one-hop lookup
\citep{guo2025lightragsimplefastretrievalaugmented}; and PathRAG selects paths
by a heuristic computed once over a precomputed path index, with no mechanism
for noticing a wrong turn \citep{chen2025pathragpruninggraphbasedretrieval}.
Agentic methods loosen the fixed strategy but not the coupling: LLM-guided
traversal over graphs
\citep{sun2024thinkongraphdeepresponsiblereasoning,luo2024reasoninggraphsfaithfulinterpretable,jiang2023structgptgeneralframeworklarge,jin2024graphchainofthoughtaugmentinglarge}
proceeds against node and edge semantics known in advance, and the BYOKG line
\citep{mavromatis2025byokgragmultistrategygraphretrieval,agarwal-etal-2024-bring} queries unfamiliar graphs but
pairs exploration with a per-graph preparation or linking stage rather than
deriving the semantics at query time (\S\ref{sec:related}).
 
What these strategies share is not a common weakness but a common coupling:
because each encodes assumptions about the graph it runs on, retrieval and
construction cannot be varied independently. Agentic methods that let a language
model walk a graph directly
\citep{sun2024thinkongraphdeepresponsiblereasoning,luo2024reasoninggraphsfaithfulinterpretable,jiang2023structgptgeneralframeworklarge,jin2024graphchainofthoughtaugmentinglarge}
loosen the fixed-strategy assumption but not the coupling: they traverse
using node and edge semantics supplied in advance, most often those of a
curated graph with a fixed ontology (\S\ref{sec:related}). The strategy we
describe is specified independently of construction: one retrieval procedure,
applied unchanged to graphs that differ in edge semantics and connectivity.
 
ArcRAG separates graph retrieval into exploration and exploitation. It reads each graph's schema by introspection, which says how
the graph is stored, and an \emph{Explorer} queries the graph at inference
time for a \emph{query-local graph interface}: the relation descriptors and
paths reachable from the entities the question names. Exploration is also prompted to surface aliases and paraphrased relation
descriptors near those entities, so later stages work over the variants a
noisy graph contains without an offline canonicalization pass. A \emph{Planner} and a
\emph{Generator} then exploit what was found, choosing paths and executing
them as Cypher, and the search returns to exploration when the evidence
falls short. No step requires graph-specific prompts, relation names, or
retuning (\S\ref{sec:harness}).

\newcommand{\yes}{\ding{51}}
\newcommand{\no}{\ding{55}}
\newcommand{\half}{$\circ$}

\begin{table*}[t]
\centering
\small
\setlength{\tabcolsep}{4pt}
\begin{tabular}{@{}l l c c c c@{}}
\toprule
System & Semantics & Prep. & Interface & Programs & Cross-constr. \\
\midrule
GraphRAG \citep{edge2024graphrag}            & own & \yes & \no & \no & \no \\
HippoRAG \citep{hipporag}                    & own & \yes & \no & \no & \half \\
HippoRAG~2 \citep{hipporag2}                 & own & \yes & \no & \no & \no \\
LightRAG \citep{guo2025lightragsimplefastretrievalaugmented} & own & \yes & \no & \no & \no \\
PathRAG \citep{chen2025pathragpruninggraphbasedretrieval}    & own & \yes & \no & \no & \no \\
StepChain GraphRAG \citep{ni2025stepchaingraphragreasoningknowledge} & own & \yes & \no & \no & \no \\
Graph-R1 \citep{luo2026graphr1agenticgraphragframework}      & own & \yes & \no & \no & \no \\
\midrule
KGP \citep{wang2023knowledgegraphpromptingmultidocument}                      & own & \yes & \no & \no & \half \\
Think-on-Graph \citep{sun2024thinkongraphdeepresponsiblereasoning} & curated & \no & \half & \no & \half \\
RoG \citep{luo2024reasoninggraphsfaithfulinterpretable}      & curated & \yes & \no & \no & \no \\
Graph-CoT \citep{jin2024graphchainofthoughtaugmentinglarge}  & prompt & \half & \no & \no & \no \\
\midrule
SymKGQA \citep{agarwal-etal-2024-symkgqa}    & curated & \half & \no & \yes & \no \\
BYOKG \citep{agarwal-etal-2024-bring}        & explored & \yes & \no & \yes & \no \\
BYOKG-RAG \citep{mavromatis2025byokgragmultistrategygraphretrieval} & schema & \half & \no & \yes & \no \\
\midrule
\textbf{ArcRAG (ours)}                       & induced & \half & \yes & \yes & \yes \\
\bottomrule
\end{tabular}
\caption{Positioning along the portability axis, to our reading of each
system's paper and released implementation. 
\textbf{Semantics}: where the retriever's knowledge of node and relation types
comes from (\emph{own}: fixed by the system's own construction;
\emph{curated}: a curated KG such as Freebase or Wikidata; \emph{prompt}: a
graph description in the prompt; \emph{explored}: offline exploration of the
target graph; \emph{schema}: a global schema read from the graph;
\emph{induced}: induced per question at inference time).
\textbf{Prep.}: a graph-specific step before questions are answered.
\textbf{Interface}: an explicit per-query description of relevant structure
passed to later stages. \textbf{Programs}: executes a graph query language.
\textbf{Cross-constr.}: the same method evaluated on more than one
construction of the same data.
\yes{}~yes, \no{}~no, \half{}~partial; partial entries are discussed in
\S\ref{sec:related}.}
\label{tab:positioning}
\end{table*}

Two questions organize the paper, each
paired with the contribution that addresses it.
 
\begin{itemize}
  \item \textbf{RQ1, mechanism.} Does inducing a query-local graph interface
  and planning executable traversals against it improve compositional
  retrieval over direct Cypher generation, and how do exploration, planning,
  and retry and pass budgets account for the difference?
  \item \textbf{RQ2, portability.} Can the same retrieval procedure, with one
  configuration per dataset and no graph-specific prompts or relation names,
  operate across knowledge graphs produced by structurally different
  construction methods, and what accounts for the accuracy that changes
  between them?
\end{itemize}
 
Our contributions are the following.
\begin{itemize}
  \item A formulation of \emph{cross-construction portability} for graph
  retrieval: whether one retrieval procedure can operate when relation
  semantics, typing, and provenance are not fixed by its construction
  pipeline (\S\ref{sec:harness}).
  \item \textsc{ArcRAG}, which induces a query-local graph interface from the
  target graph at inference time and plans, executes, and repairs
  compositional traversals against it (\S\ref{sec:harness}).
  \item A controlled evaluation: a comparison with a Direct-Cypher agent,
  component ablations across retry and pass budgets on both graph families,
  and a paired transfer design over schema-guided and open-vocabulary
  constructions of the same three corpora (\S\ref{sec:experiments},
  \S\ref{sec:results}).
\end{itemize}

Within the two LLM-extracted, entity-centric constructions we evaluate, the
results support a bounded form of cross-construction portability: retrieval
need not be co-designed with each construction when the retriever operates
against an interface induced from the graph at query time.

\section{Related Work}
\label{sec:related}

\paragraph{GraphRAG over a fixed construction.}
GraphRAG \citep{edge2024graphrag} precomputes community summaries over a
hierarchical partition; HippoRAG \citep{hipporag} runs personalized PageRank
over an OpenIE graph, and HippoRAG~2 \citep{hipporag2} adds denser phrase and
passage indexing; LightRAG \citep{guo2025lightragsimplefastretrievalaugmented}
maintains entity and relation keyword indices; and PathRAG
\citep{chen2025pathragpruninggraphbasedretrieval} prunes relational paths. A
second group varies what is indexed: hyperedges for $n$-ary facts
\citep{luo2025hypergraphragretrievalaugmentedgenerationhypergraphstructured},
hierarchical aggregation
\citep{huang2025retrievalaugmentedgenerationhierarchicalknowledge},
token-efficient construction
\citep{xiao2025teragtokenefficientgraphbasedretrievalaugmented}, separate
similarity and relatedness indices \citep{zhang2025sireragindexingsimilarrelated},
context-dependent structure \citep{lau2026breakingstaticgraphcontextaware}, and
path tracking \citep{li2025neuropathneurobiologyinspiredpathtracking}. In each
case retrieval is designed together with the structure its own indexing step
produces. The hierarchical lexical graph \citep{Ghassel_2025} makes this
coupling explicit: its retrievers address the statement, fact, topic, and
chunk node types of its own graph model directly in their graph queries.
Some works do vary construction: KGP
\citep{wang2023knowledgegraphpromptingmultidocument} compares several
passage-similarity graphs, HippoRAG ablates its extractor, and
\citet{han2026ragvsgraphragsystematic} find GraphRAG performance stable across the
LLMs used for construction. Studies that compare constructions directly hold
a simple retriever fixed in order to judge the graphs
\citep{dacruz2025ontologylearningknowledgegraph}, or recombine fixed retrieval operators with
different graph types to find the best pairing \citep{zhou2026indepthanalysisgraphbasedrag}. In
each case the graph is the object of study and the retriever an instrument;
whether a system transfers across constructions that differ in schema
commitment, typing, and provenance is not tested.
Surveys characterize when graph structure helps over flat retrieval
\citep{xiang2026usegraphsragcomprehensive}. Table~\ref{tab:positioning}
summarizes these differences.

\paragraph{Iterative and agentic graph retrieval.}
IRCoT \citep{trivedi2023interleavingretrievalchainofthoughtreasoning}
interleaves reasoning with passage retrieval, and KGP traverses its passage
graph with a fine-tuned LLM. Over knowledge graphs, StructGPT
\citep{jiang2023structgptgeneralframeworklarge} exposes reading interfaces for
structured data; Think-on-Graph
\citep{sun2024thinkongraphdeepresponsiblereasoning} extends reasoning paths by
querying candidate relations and entities step by step; RoG
\citep{luo2024reasoninggraphsfaithfulinterpretable} plans relation paths and
grounds them in the graph; Graph-CoT
\citep{jin2024graphchainofthoughtaugmentinglarge} interleaves reasoning with
graph function calls; StepChain GraphRAG
\citep{ni2025stepchaingraphragreasoningknowledge} decomposes questions and
searches a graph built on the fly from retrieved passages; and Graph-R1
\citep{luo2026graphr1agenticgraphragframework} trains an agentic retriever with
reinforcement learning. Several of these explore the graph at query time, so
we do not claim novelty for LLM-guided traversal. They differ from ArcRAG in
two respects. Exploration is interleaved step by step with answering rather
than summarized into an interface that a planner commits to. And each is
developed and evaluated on a single graph, most often a curated KG whose
relation names are meaningful by design; Think-on-Graph also compares source
KGs on the same questions, but each is a curated ontology rather than a
construction of the corpus.

\paragraph{Querying unfamiliar graphs.}
Closest to our setting is work that answers questions over graphs not seen
during development. BYOKG \citep{agarwal-etal-2024-bring} explores an unseen
graph once to generate query--program exemplars that ground later program
synthesis; SymKGQA \citep{agarwal-etal-2024-symkgqa} generates step-by-step
symbolic programs from few-shot examples and grounds them with a
retrieval-based resolver; and text-to-Cypher work
\citep{ozsoy2024text2cypherbridgingnaturallanguage} translates
questions into graph queries given a schema.
GRA \citep{dragic2026schemaagnosticgraphreasoningagent} navigates a hybrid knowledge graph with generic
tools and discovers everything domain-specific at run time, but is evaluated
on a single graph.
The most similar system,
BYOKG-RAG \citep{mavromatis2025byokgragmultistrategygraphretrieval}, generates
entities, relation paths, and OpenCypher queries over custom graphs and
refines them against retrieved context. It makes no construction assumptions,
but its retrievers presuppose the conventions of curated KGs: relation
semantics carried by edge types listed in a global schema, nodes addressed
by name, and meaningful edge direction. Graphs whose relations are
open-vocabulary strings stored as edge properties, as in our
open-information-extraction constructions, violate these conventions
: their schema exposes a single relationship type, so retrievers that select
among types have nothing to select. This is the setting a query-local
interface is designed for. These systems are
also evaluated as KGQA over curated benchmark graphs (Freebase-based GrailQA
and MetaQA for BYOKG; encyclopedic, temporal, and medical KGs for BYOKG-RAG),
where answers are graph entities; none evaluates multiple constructions of
the same corpus.They establish that LLMs can synthesize executable programs over graphs they
were not built for, and ArcRAG builds on that; our Direct-Cypher agent
(\S\ref{sec:experiments}) represents this approach under our setup.

\section{Knowledge Graph Construction}
\label{sec:kg-construction}

\begin{figure*}[t]
\centering
\includegraphics[width=\textwidth,height=0.5\textheight,keepaspectratio]{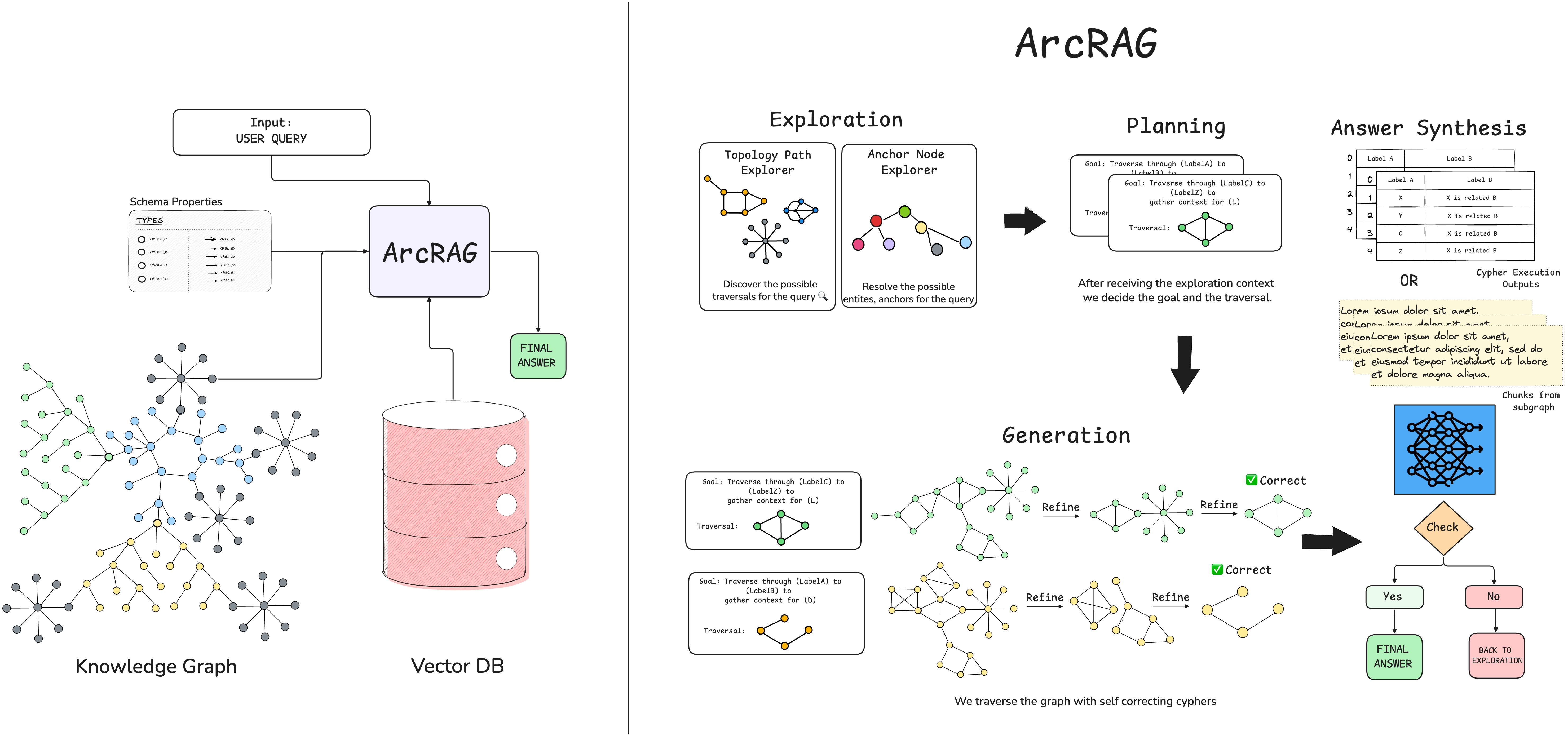}
\caption{\textbf{ArcRAG} at inference (left) and the harness (right): exploration
$\rightarrow$ planning $\rightarrow$ generation $\rightarrow$ synthesis.}
\label{fig:arch}
\end{figure*}

\begin{table}[t]
\centering
\small
\setlength{\tabcolsep}{6pt}
\begin{tabular}{@{}llrr@{}}
\toprule
Graph & Corpus & Nodes & Edges \\
\midrule
FinReflectKG    & MuSiQue  &  58{,}418 &  80{,}676 \\
                & HotpotQA &  33{,}203 &  63{,}493 \\
                & TriviaQA &  72{,}266 &  99{,}915 \\
\addlinespace
\midrule
\addlinespace
HippoRAG OpenIE & MuSiQue  &  85{,}520 & 142{,}315 \\
                & HotpotQA & 101{,}200 & 130{,}053 \\
                & TriviaQA & 162{,}687 & 193{,}470 \\
\bottomrule
\end{tabular}
\caption{Graphs produced by the two construction methodologies over the same
three corpora.}
\label{tab:kg-stats}
\end{table}

We construct graphs over the same corpora using two extraction
methodologies, chosen because they differ along three of the axes identified
in \S\ref{sec:intro}: schema commitment, relation vocabulary, and
connectivity. Both are existing published pipelines rather than
contributions of this work, applied unmodified except where a corpus
requires it, and neither is presupposed by the retrieval harness. The order below is expositional and implies no
precedence.

\paragraph{Schema-guided construction.}
We apply FinReflectKG's pipeline
\citep{arun2025finreflectkgagenticconstructionevaluation}, in which an
extraction agent proposes triples chunk by chunk and a reflection pass
checks each proposal for faithfulness to its source chunk before committing
it, to MuSiQue, HotpotQA, and TriviaQA. Two adaptations are inherently
corpus-specific. In place of table-aware chunking we chunk at the passage
level, following each corpus's existing paragraph segmentation so that chunk
boundaries align with the units the hybrid dense and BM25 retriever already
indexes. In place of the business-defined disclosure schema --- segments,
risk factors, related parties --- we extract against an LLM-constructed
schema suited to encyclopedic text. The extraction-and-commit reflection
loop is otherwise the same loop used for the financial 10-K graph. The
resulting graphs have typed nodes and a closed relation set.

\paragraph{Open-vocabulary construction.}
Over the same chunks we build a second set of graphs using HippoRAG's
methodology \citep{hipporag}: open information extraction with
no predefined relation vocabulary, entity nodes linked across passages by
embedding-based synonym edges, and the graph treated as a corpus-wide
superposition rather than a per-document structure.

\paragraph{What the harness consumes.}
Both families are stored as labeled property graphs; OpenIE holds its
predicates in an edge property under a single relationship type. The harness
reads each graph's schema by introspection, with no hand-written schema or
construction-specific names. The one other difference it observes, whether
edges point to source chunks, selects the retrieval store and the evidence
answers are composed from (\S\ref{sec:synthesis}).

\begin{algorithm}[!t]
\footnotesize
\caption{ArcRAG retrieval}
\label{alg:arcrag}
\begin{algorithmic}[1]
\REQUIRE question $q$, graph $G$, accompanying store $\mathcal{D}$
\ENSURE answer $a$
\STATE $q' \gets q$, \; $W \gets \emptyset$
\REPEAT
  \STATE $\mathcal{R} \gets \textsc{Retrieve}(\mathcal{D}, \textsc{Phrases}(q'))$
  \STATE $A \gets \textsc{Anchors}(\mathcal{R})$
  \IF{$\mathcal{D}$ is a chunk store}
    \STATE $G_q \gets G[V_{\mathcal{R}}]$, \; $\Sigma \gets \emptyset$
  \ELSE
    \STATE $G_q \gets G$, \; $\Sigma \gets \mathcal{R}$ \COMMENT{triples as hints}
  \ENDIF
  \STATE $\hat{S}_q, \mathcal{I}_q \gets \textsc{Explore}(G_q, A, \Sigma)$
  \STATE $P \gets \textsc{Plan}(q', \hat{S}_q, \mathcal{I}_q)$
  \FOR{each subplan $\pi \in P$}
    \STATE $B \gets A$
    \FOR{each hop $h \in \pi$ in order}
      \STATE $Q \gets \textsc{Compile}(h, \hat{S}_q, B)$; \; $R_h \gets \textsc{Exec}(Q)$
      \WHILE{$\neg\textsc{Acceptable}(R_h)$ within budget}
        \STATE $Q \gets \textsc{Refine}(Q, R_h)$; \; $R_h \gets \textsc{Exec}(Q)$
      \ENDWHILE
      \STATE \textbf{if} $\neg\textsc{Acceptable}(R_h)$ \textbf{then break}
      \STATE $B \gets \textsc{Nodes}(R_h)$; \; $W \gets W \cup \textsc{Edges}(R_h)$
    \ENDFOR
  \ENDFOR
  \STATE $Z \gets$ \textsc{Passages}$(W)$ if chunk store, else \textsc{Rows}$(W)$
  \STATE $a \gets \textsc{Compose}(q, Z)$
  \STATE $\mathit{ok}, q' \gets \textsc{Sufficient}(q, a, Z)$
\UNTIL{$\mathit{ok}$ \OR $q' = \bot$ \OR iteration cap}
\RETURN $a$
\end{algorithmic}
\end{algorithm}

\begin{table}[t]
\centering
\footnotesize
\setlength{\tabcolsep}{3pt}
\begin{tabular}{@{}p{0.24\columnwidth}p{0.35\columnwidth}p{0.33\columnwidth}@{}}
\toprule
Function & Fixed in code & Decided by the LLM \\
\midrule
\textsc{Explore} & read-only queries; explorer, turn, and depth budget & which relations and paths to follow \\
\textsc{Plan} & output format; [validation against $\hat S_q$] & subplans, hop order, answer type \\
\textsc{Acceptable} & error, empty result, unbound variable, row cap & [none] \\
\textsc{Refine} & traversal shape fixed; attempt budget & revised predicates and bindings \\
\textsc{Sufficient} & iteration cap & whether evidence answers $q$; follow-up $q'$ \\
\bottomrule
\end{tabular}
\caption{What each function of Algorithm~\ref{alg:arcrag} fixes in code and
what the LLM decides.}
\label{tab:predicates}
\end{table}

\section{Methodology}
\label{sec:harness}

Real-world knowledge graphs are incomplete and noisy
\citep{10.3233/SW-160218}: one entity is stored under several names, one
relation as several predicate strings, and edges a question needs may be
missing; open information extraction leaves entity and relation phrases
uncanonicalized altogether \citep{Vashishth_2018}. A retriever written
against one construction's schema inherits its assumptions about all three.
ArcRAG's stages can instead be read through the exploration exploitation
trade-off familiar from search \citep{kocsis2006ecml-bandit}, although no
stage implements a bandit or tree-search policy. The Explorer explores: it establishes what the graph
contains around the entities a question names, including their variants,
without committing to an answer. The Planner and Generator exploit: they
commit to the most promising explored paths and execute them as queries.
When the evidence does not support an answer, the search returns to
exploration. The budgets reflect the same trade-off: retries extend exploitation and
passes renew exploration, and how much the Planner's guidance adds depends
on both the budget and the graph (Appendix~\ref{app:component-ablations}).

\subsection{Setup}
\label{sec:setup}
The target is a labeled property graph $G=(V,E,\ell_V,\ell_E,p_V,p_E)$ with
node and edge labels and key--value properties. Its schema $S_G$, the node
labels, relationship types, and property keys, is read by introspection
before any question (line~1); none is written by hand. $S_G$ shows where
relations are stored: in relationship types for a schema-guided graph, in an
edge property for an open-vocabulary one. We write $\rho(e)$ for the
resulting \emph{relation descriptor} of edge $e$. For a question $q$ with
anchors $A\subseteq V$, the entities it names, exploration yields the
\emph{query-local interface} $\hat S_q$: the descriptors and signatures
$(t_i,\rho,t_j)$ observed near $A$, with the paths $\mathcal I_q$ that
confirm them. Executed queries return the \emph{witness} $W$, the subgraph
connecting the anchors to an answer. Only provenance depends on the graph
family: whether edges point to source chunks selects the store that seeds
retrieval and the evidence the answer reads (lines~6--10 and~20). Beyond
introspection, ArcRAG needs only a hybrid text index over that store.
Algorithm~\ref{alg:arcrag} and Table~\ref{tab:predicates} specify the
procedure.

\subsection{Explore}
\label{sec:explore}
Without exploration, a query generator guesses which entities and relations
exist: $S_G$ says how the graph is stored, not what lies near a question.
Exploration supplies this, and canonicalizes at query time: explorers are
prompted to surface candidate aliases and paraphrased relation descriptors
near the anchors, so later stages reason over the observed variants instead
of relying on an offline canonicalization pass. Phrases from the
question are retrieved by reciprocal rank fusion
\citep{cormack2009reciprocal} of dense and sparse search (line~4) to obtain
anchors (line~5); chunks bound later queries to the subgraph they induce,
while retrieved triples only hint where to look (lines~6--10). Two explorers
each issue one batch of read-only Cypher queries (line~11), locating
relations through $S_G$. Their report, $\hat S_q$, scales with the
exploration budget, not the graph, and is a per-question view, not an
ontology. Exploration gives the largest gain in our ablations
(Appendix~\ref{app:component-ablations}).

\subsection{Plan}
\label{sec:plan}
The Planner is the exploitation heuristic: it decides which explored paths
are worth executing, so the Generator need not find them by trial and error.
From $S_G$, $\hat S_q$, and $\mathcal I_q$ it fixes the answer type and emits
a bounded set of subplans, each a traversal from anchors it attaches
(line~12); a bridge question needs one, a comparison or aggregation several.
One subplan is always an open neighborhood scan, so some evidence is gathered
even if targeted subplans fail. The plan is fixed for its pass. Its value depends on the graph and the budget: on the typed graph it helps
most when retries are scarce, while on the open-vocabulary graph it is
neutral at small budgets and slightly ahead at large ones
(Appendix~\ref{app:component-ablations}).

\subsection{Generate}
\label{sec:generate}
The Generator executes rather than proposes, so every result can be checked
before it reaches the answer. It compiles each subplan into one read-only
Cypher query over the descriptors and variants surfaced in $\hat S_q$
(line~14). One query chains the subplan's hops: an open far end enumerates a
set, the next hop steps across the rows returned, and a filter applies a
constraint no single passage states. On chunk-backed graphs, queries are
scoped to the retrieved chunks. The first attempt is the least constrained;
an unacceptable result (Table~\ref{tab:predicates}) is refined by revising
predicates and bindings while the traversal shape stays fixed
(lines~15--17). This limits uncontrolled rewrites but does not guarantee that
a correct earlier candidate survives. A subplan still unacceptable when its
budget runs out is left to the outer loop; acceptable results join the
witness (line~18).

\subsection{Answer Synthesis}
\label{sec:synthesis}
The witness determines what the answer model reads (lines~20--21). Where its
edges point to source chunks, the answer is composed from those passages,
which keep qualifications a triple drops; otherwise it is composed from the
returned rows, at the coarser grain of an edge. The graph's provenance, not a
per-experiment choice, decides which applies. Each answer can be traced to
the evidence it was given and the witness re-executed, though we do not
verify claim-level attribution.

\subsection{Retrieve--Answer Loop}
\label{sec:loop}
The loop is where the search returns to exploration. Within a pass, an
unacceptable result triggers another attempt; across passes, an answer the
evidence does not support yields a follow-up question $q'$ that seeds a fresh
pass (lines~22--23), which can reach relations or paths the previous one
missed. The witness accumulates across passes, and the loop stops when the
answer is judged sufficient, no follow-up can be formed, or the cap is
reached. More passes raise accuracy in every ablated configuration
(Appendix~\ref{app:component-ablations}).

\begin{table*}[t]
\centering
\small
\begin{tabular}{lccccc}
\toprule
& \multicolumn{2}{c}{MuSiQue} & \multicolumn{2}{c}{HotpotQA} & TriviaQA \\
\cmidrule(lr){2-3}\cmidrule(lr){4-5}\cmidrule(lr){6-6}
Method & EM & F1 & EM & F1 & F1 \\
\midrule
GraphRAG~\citep{edge2024graphrag}          & 27.3 & 38.5 & 55.2 & 68.6 & 45.1 \\
LightRAG~\citep{guo2025lightragsimplefastretrievalaugmented}           & 20.0 & 29.3 & 33.3 & 44.9 & 40.2 \\
PathRAG~\citep{chen2025pathragpruninggraphbasedretrieval}            &  8.4 & 20.8 & 23.8 & 41.2 & 37.4 \\
HippoRAG~\citep{hipporag}     & 27.8 & 40.5 & 58.5 & 73.1 & ---  \\
HippoRAG 2~\citep{hipporag2}  & 37.2 & 48.6 & 62.7 & 75.5 & 46.9 \\
HyperGraphRAG~\citep{luo2025hypergraphragretrievalaugmentedgenerationhypergraphstructured} & 10.1 & 23.2 & 29.5 & 46.9 & 45.0 \\
HiRAG~\citep{huang2025retrievalaugmentedgenerationhierarchicalknowledge}               & ---  & ---  & 37.0 & 52.3 & ---  \\
TERAG~\citep{xiao2025teragtokenefficientgraphbasedretrievalaugmented}                    & 18.8 & 29.6 & 46.9 & 59.8 & ---  \\
CatRAG~\citep{lau2026breakingstaticgraphcontextaware}                  & ---  & 45.0 & ---  & ---  & ---  \\
NeuroPath~\citep{li2025neuropathneurobiologyinspiredpathtracking}            & 31.4 & 44.3 & 50.5 & 64.7 & ---  \\
StepChain GraphRAG~\citep{ni2025stepchaingraphragreasoningknowledge} & ---  & ---  & 66.7 & 79.5 & ---  \\
Graph-R1~\citep{luo2026graphr1agenticgraphragframework}            & ---  & 46.2 & ---  & 62.7 & 71.9 \\
\midrule
\textbf{ArcRAG (w/ FinReflectKG)} & \textbf{44.0} & \textbf{53.0} & \textbf{74.5} & \textbf{80.8} & \textbf{87.9} \\
\bottomrule
\end{tabular}
\caption{Best EM and F1 reported for each method on each dataset, taken
across all models and configurations its authors evaluated. Protocols,
answer models, and scoring code differ across sources, so this is a
best-reported summary rather than a controlled comparison. Prior work
reports F1 only on TriviaQA.}
\label{tab:best-published}
\end{table*}

\begin{table}[t]
\centering
\footnotesize
\setlength{\tabcolsep}{4pt}
\resizebox{\columnwidth}{!}{%
\begin{tabular}{lccccc}
\toprule
& \multicolumn{2}{c}{MuSiQue} & \multicolumn{2}{c}{HotpotQA} & TriviaQA \\
\cmidrule(lr){2-3}\cmidrule(lr){4-5}\cmidrule(lr){6-6}
Method & EM & F1 & EM & F1 & F1 \\
\midrule
HippoRAG   & 26.2 & 35.1 & 52.6 & 63.5 & ---  \\
HippoRAG 2 & 37.2 & 48.6 & 62.7 & 75.5 & 46.9 \\
\addlinespace
ArcRAG (w/ OpenIE Graph) & 38.0 & 45.0 & 65.5 & 72.2 & 71.6 \\
\textbf{ArcRAG (w/ FinReflectKG)}   & \textbf{44.0} & \textbf{53.0} & \textbf{74.5} & \textbf{80.8} & \textbf{87.9} \\
\bottomrule
\end{tabular}}
\caption{Transfer across knowledge-graph families. All ArcRAG prompts, models, and hyperparameters are fixed within a dataset;
only the graph family and its provenance-determined evidence store change. HippoRAG and HippoRAG~2 figures for MuSiQue and
HotpotQA are as reported by
\citet{bai2025autoschemakgautonomousknowledgegraph} under a common harness;
the HippoRAG~2 TriviaQA figure is from
\citet{luo2026graphr1agenticgraphragframework}. As above, prior work reports
F1 only on TriviaQA.}
\label{tab:ablation}
\end{table}

\section{Experimental Setup}
\label{sec:experiments}

\subsection{Datasets and Samples}
We evaluate on three QA benchmarks that differ in how much composition an
answer requires. \textbf{MuSiQue}~\citep{trivedi2022musiquemultihopquestionssinglehop}
chains up to four single-hop questions, with the bridging entities absent
from the question text. \textbf{HotpotQA}~\citep{yang2018hotpotqadatasetdiverseexplainable}
is predominantly two-hop. \textbf{TriviaQA}~\citep{joshi2017triviaqalargescaledistantly}
is predominantly single-hop and serves as a control for whether the harness
degrades performance when little traversal is needed. For each dataset, both
graph families are built and frozen before evaluation
(Table~\ref{tab:kg-stats}), and questions are answered end to end, with no
gold passages, entities, or hop counts supplied. Main results follow the
validation-sample convention of HippoRAG and IRCoT~\citep{hipporag}. The component ablations use a stratified sample of 100 questions drawn from
all three datasets and run on both graph families.

\subsection{Baselines}
We compare against two kinds of reference. The first is results reported
for prior GraphRAG systems (Table~\ref{tab:best-published}), most of which
design retrieval together with their own graph construction
(\S\ref{sec:related}). The second is a Direct-Cypher agent: the Generator
alone with a single attempt and a single pass, compiling Cypher from the
question and the introspected schema with no exploration, planning,
refinement, or follow-up pass. Given more retries and passes, the Generator
alone becomes an ablation of ArcRAG, which we compare with Explorer +
Generator and the full pipeline at each budget.

\subsection{Protocol}
Within each dataset, the configuration is identical for both graph families;
only the graph and its accompanying store change, as determined by provenance
(\S\ref{sec:setup}). For the transfer comparison
(Table~\ref{tab:ablation}), baseline figures come from the common harness of
\citet{bai2025autoschemakgautonomousknowledgegraph}. The external systems
answer with Llama-3.3-70B-Instruct, GPT-4, or Qwen2.5 models depending on
the source, whereas ArcRAG uses gpt-oss-120b for its agents and
Nemotron-3-Super-120B-A12B for answering (Appendix~\ref{app:hyperparams});
these comparisons share a benchmark convention but not a backbone, and we
treat them as context rather than as controlled retriever comparisons.

\subsection{Metrics}
Exact match (EM) is the primary metric; token-level F1 is reported for
comparability with prior work, which on TriviaQA reports F1 only. Both use
the normalization of prior GraphRAG work~\citep{hipporag}, taking the maximum
over references and averaging over questions. Ablations report EM. Token
footprints are given in Appendix~\ref{app:tokencost}.

\subsection{Configuration}
\label{sec:config}
Models, budgets, caps, and retrieval settings are listed in
Appendix~\ref{app:hyperparams}. Within each dataset, every setting is the
same for both graph families.

\section{Results}
\label{sec:results}

Studies that vary graph construction hold a simple retriever fixed to judge
the graphs, and compare graphs from a single construction family
\citep{dacruz2025ontologylearningknowledgegraph}. We ask the reverse
question: whether one system, with its configuration fixed, stays effective
when a schema-guided construction is replaced by an open-vocabulary one.
We report what each stage contributes across retry and pass budgets (RQ1),
how accuracy carries across the two constructions (RQ2), and where the
results sit among reported figures.

\subsection{Mechanism}
\label{sec:mechanism}
For RQ1, query-local exploration improves on direct Cypher generation at
every budget we test: on the stratified ablation sample, adding the Explorer
raises accuracy at all 18 retry and pass settings, by 15--39 points on
FinReflectKG and 12--32 on the open-vocabulary graphs
(Appendix~\ref{app:component-ablations}). At five retries and five passes,
ArcRAG's best configurations reach 88.0\% and 74.0\%, against 50.0\% and
59.0\% for the Generator alone and 12.0\% and 6.0\% for the single-attempt
Direct-Cypher agent. Planning is graph-dependent: it gives the best accuracy
at budgets up to 9 on FinReflectKG, but not on the open-vocabulary graphs.
Because exploration and planning add LLM calls, these comparisons control
retries and passes, not total inference cost.

\subsection{Cross-Construction Transfer}
\label{sec:transfer}
The open-vocabulary graphs carry one node label, one relationship type, and
no chunk provenance (\S\ref{sec:kg-construction}). With the same
configuration, ArcRAG obtains 38.0 EM on MuSiQue, 65.5 EM on HotpotQA, and
71.6 F1 on TriviaQA over them (Table~\ref{tab:ablation}). Their introspected
schema is nearly empty, so the relation semantics the system uses come from
the query-local interface alone. The schema-guided graphs score higher by
6.0 EM, 9.0 EM, and 16.3 F1: portability is not invariance, and because the
two families differ in both structure and provenance, these gaps reflect
their combined effect. Against HippoRAG and HippoRAG~2 as reported by
\citet{bai2025autoschemakgautonomousknowledgegraph}, the open-vocabulary row
is well above HippoRAG and within 4 points of HippoRAG~2 on MuSiQue and
HotpotQA, higher on EM and lower on F1. For RQ2, one configuration operates
across both constructions and remains competitive on the one with no
relation schema.

\subsection{Position Against Reported Results}
\label{sec:reported}
Table~\ref{tab:best-published} lists, for each prior system, the best figure
its authors report on each dataset, taken across all models and
configurations they evaluated. With the schema-guided graph, ArcRAG obtains
44.0 EM on MuSiQue, 74.5 EM on HotpotQA, and 87.9 F1 on TriviaQA, above every
figure in the table on every metric; the narrowest margin is HotpotQA F1,
80.8 against 79.5 for StepChain GraphRAG. Because protocols, answer models,
and scoring code differ across sources, this places ArcRAG at the top of the
reported range rather than establishing a controlled improvement over any
one system.

\section*{Limitations}
This is an early study of cross-construction portability, and its scope
reflects that. We evaluate two graph families over the same three corpora,
chosen to differ in kind rather than degree: schema-guided extraction with
typed nodes, a closed relation set, and chunk provenance, against open
information extraction with untyped nodes, free-text relations, and no
provenance. Two maximally different constructions test portability more
severely than several similar ones would, but they do not establish transfer
to curated ontologies or to graphs with different node granularities.

Because the two families differ in both structure and provenance, the
accuracy gap between them reflects both, and we do not separate the two.
Canonicalization at query time is a design property of the harness; we do
not measure how much noise or variation it absorbs. Comparisons with prior
GraphRAG systems use their best reported figures, whose protocols and answer
models differ from ours, rather than reruns under a shared setup. Each
configuration is run once, and the answer model samples at temperature 1.0
and also judges whether its own evidence is sufficient, so the results carry
run-to-run variance that we do not quantify.

The benchmarks are public and the answer model is a large pretrained model,
so parametric knowledge may contribute to absolute scores; comparisons that
share the answer model limit but do not remove this effect. Our experiments
use encyclopedic QA and do not establish reliability for the regulated
domains that motivate traceable retrieval: an executable witness does not
guarantee factual correctness, and deployment there would require
domain-specific validation and human oversight.

Finally, graph construction still affects accuracy, so we do not claim that
retrieval and construction never benefit from co-design.

\section*{Ethical Considerations}
\paragraph{Data.} All experiments use public benchmarks built from Wikipedia
and web text: MuSiQue (CC BY 4.0), HotpotQA (CC BY-SA 4.0), and TriviaQA (CC BY 4.0). We use them for research evaluation, consistent with their
licenses, and the knowledge graphs we build from them are derived artifacts
of the same text. These corpora concern public figures and encyclopedic
facts; we collect no new data from people and process no private
information.

\paragraph{Models.} Agents use gpt-oss-120b (Apache 2.0) and answers use
Nemotron-3-Super-120B-A12B (NVIDIA Nemotron Open Model License); dense
retrieval uses Qwen3-Embedding-0.6B (Apache 2.0). All are openly released
models, used within the terms of their licenses. Graphs extracted by
language models inherit their errors and biases, as well as those of the
source text, and retrieval over such a graph can surface them.

\paragraph{Intended use and risks.} Our motivation includes regulated
domains where answers must be traceable, but our experiments use
encyclopedic QA and do not establish reliability there. An executable
witness shows how an answer was reached, not that it is correct, and we do
not verify claim-level attribution. Any deployment in finance, law,
healthcare, or compliance would require domain-specific validation and human
oversight.

\bibliography{aaai2027}
\twocolumn[{\centering\Large\bfseries Appendix\par\vspace{1.5em}}]
\appendix

\section{Overview}
\label{app:overview}
These appendices give implementation details and hyperparameters
(Appendix~\ref{app:impl}), a token-cost analysis including a comparison by
graph family (Appendix~\ref{app:tokencost}), full component-ablation results
(Appendix~\ref{app:component-ablations}), a qualitative example
(Appendix~\ref{app:qualitative}), and reproducibility notes
(Appendix~\ref{app:repro}).

\begin{table*}[!t]
\centering
\small
\begin{tabular}{lcccc}
\toprule
Stage & Temp. & Top-$p$ & Thinking & Max tokens \\
\midrule
Query refinement            & 0.0 & 0.95 & off & 8{,}192 \\
Search-phrase extraction    & 0.0 & 0.95 & off & 8{,}192 \\
Explorer Cypher generation  & 0.1 & 0.95 & off & 8{,}192 \\
Explorer report synthesis   & 0.2 & 0.95 & off & 8{,}192 \\
RootPlanner                 & 0.0 & 0.95 & off & 8{,}192 \\
Planner subplan expansion   & 0.2 & 0.95 & off & 8{,}192 \\
Generator (Cypher)          & 0.2 & 0.95 & off & 8{,}192 \\
\textbf{Answer + sufficiency} & \textbf{1.0} & \textbf{0.95} & \textbf{on} & \textbf{16{,}000} \\
\bottomrule
\end{tabular}
\caption{Sampling parameters by pipeline stage.}
\label{tab:sampling}
\end{table*}
The reasoner (gap analysis) makes \emph{no} LLM call --- its output is
built deterministically from the retrieved triplets.

\subsubsection{Stage Settings}
\begin{table*}[!t]
\centering
\small
\begin{tabular}{@{}lp{0.62\textwidth}@{}}
\toprule
Setting & Value \\
\midrule
Parallel explorers & 2 (anchored-path, topology-path) \\
Cyphers per explorer & 6 (one shot) \\
Explorer turns & 3 (passed; unused on the active subgraph-bound path) \\
Explorer timeouts & LLM 30\,s / Cypher 5\,s; report retried once \\
RootPlanner & 2 attempts, 30\,s timeout each \\
\textbf{Subplan hard cap} & \textbf{4} ($\ge$1 is a mandatory open-ended neighborhood scan) \\
Subplan expansion & LLM-based expansion plus deterministic skeleton conversion \\
Generator attempts / subplan & 5 base $+$ up to 2 extra retries ($\le 7$ total); Cypher timeout 10\,s \\
Loop iteration cap & 6 \\
\bottomrule
\end{tabular}
\caption{Stage-level settings shared across corpora.}
\label{tab:stage-settings}
\end{table*}

\section{Additional Implementation Details}
\label{app:impl}

All values in this section reflect the pipeline's source and
configuration as run. The three corpora (MuSiQue, HotpotQA, TriviaQA) are
served by structurally identical pipelines, each built on a single shared
processing loop. The four agents --- Exploration, RootPlanner, Planner, and Generator ---
contain no construction-specific names in their prompts or code. A corpus
differs only in its per-corpus constants (Memgraph connection, Qdrant
collection, retrieval depth), its schema, which is read from the graph by
introspection rather than written by hand, and its per-corpus context
prompt.

\subsection{System Overview}
\label{app:system-overview}
Per query, each endpoint runs an iterative retrieve--answer loop
(Algorithm~\ref{alg:arcrag}), capped at 6 passes:
\begin{enumerate}
\item \textbf{Introspect} the graph's schema (node labels, relationship
types, and property keys) once per graph, and cache it.
\item \textbf{Rewrite} the query once, before entering the loop.
\item Loop, each pass running the shared pipeline:
\begin{itemize}
\item \textbf{Exploration} --- extract search phrases $\rightarrow$ hybrid
retrieve $\rightarrow$ deterministic gap analysis $\rightarrow$ 2 parallel
graph explorers.
\item \textbf{Planning} --- the RootPlanner decomposes the query into at
most 4 ordered subplans; the PlannerAgent attaches each subplan's binding
anchors.
\item \textbf{Generation} --- the GeneratorAgent emits one read-only
Cypher query per subplan.
\item Execute every subplan's final Cypher, render deduplicated chunk
context, answer the original query, and self-judge sufficiency.
\item Stop once the answer is judged sufficient, or the proposed
follow-up query is empty; otherwise re-run on the follow-up query,
accumulating context across passes.
\end{itemize}
\end{enumerate}

\emph{Terminology note.} RootPlanner and PlannerAgent together implement
the Planner stage (\textsc{Plan} in Algorithm~\ref{alg:arcrag});
GeneratorAgent implements the Generator (\textsc{Compile} and
\textsc{Refine}); the two explorers implement \textsc{Explore}; and the
answerer implements \textsc{Compose} and \textsc{Sufficient}. A
\emph{pass} is one iteration of the outer loop; \emph{hops} refer only to
traversal depth within a subplan's query. Node-creator and
visualize/report/extractor calls (Table~\ref{tab:fixed-costs}) are
internal sub-steps of Generation, not separately named stages.

\subsection{Hyperparameters}
\label{app:hyperparams}

\subsubsection{Models}
\begin{table}[!t]
\centering
\small
\setlength{\tabcolsep}{4pt}
\begin{tabular}{@{}p{0.40\columnwidth}p{0.55\columnwidth}@{}}
\toprule
Role & Model \\
\midrule
Agents (explorers, planners, generator) & gpt-oss-120b \\
Query rewrite & gpt-oss-120b \\
Answer composition and sufficiency judge & Nemotron-3-Super-120B-A12B \\
Dense embedder & Qwen3-Embedding-0.6B \\
Sparse retriever & BM25 \\
\bottomrule
\end{tabular}
\caption{Models by pipeline role. Full model identifiers are given in
Appendix~\ref{app:repro}; BM25 uses a server-side IDF modifier, and the
embedder its native width.}
\label{tab:models}
\end{table}
LLMs and embeddings are served via an OpenAI-compatible gateway.

Each graph's schema is read from Memgraph by introspection and cached;
the same schema is injected into every agent call for that graph. Each
corpus is queried through a dedicated client instance bound to its own
port.

\emph{Graph family scope.} The following applies to the FinReflectKG
(typed, chunk-backed) runs: all three are \emph{typed} chunk-KGs
(entities extracted from Wikipedia/web text; every relationship carries a
back-pointer to its source chunk). The OpenIE runs (Table~\ref{tab:persisted-footprint})
differ: open-vocabulary extraction with no fixed relation vocabulary and
no chunk back-pointers. They are stored as labeled property graphs with a
single node label and relationship type, the predicate held in an edge
property, so their introspected schema is minimal; it is injected in the
same way.

\subsubsection{Retrieval: Hybrid Dense + BM25 with RRF Fusion}
Per search phrase, a dense leg and a BM25 leg each prefetch a fixed
number of candidates, fused with weighted Reciprocal Rank Fusion,
returning the top-$k$ hits. No multivector rerank is applied.
\begin{table}[!htbp]
\centering
\small
\setlength{\tabcolsep}{4pt}
\begin{tabular}{@{}lccc@{}}
\toprule
Knob & MuSiQue & HotpotQA & TriviaQA \\
\midrule
Dense prefetch          & 30 & 30 & 30 \\
BM25 prefetch           & 30 & 30 & 30 \\
Top-$k$ fused hits & 10  & 10  & 10  \\
\bottomrule
\end{tabular}

\caption{Retrieval knobs by corpus. These are the only per-corpus retrieval
settings; within a corpus they are identical for both graph families.}
\label{tab:retrieval}
\end{table}

\begin{table}[!t]
\centering
\small
\begin{minipage}[t]{0.45\textwidth}
\centering
\begin{tabular}{lr}
\toprule
Prompt & Tokens \\
\midrule
Root planner                        & 3{,}354 \\
Generator                           & 1{,}415 \\
Explorer (anchored / topology)      & 1{,}377 / 1{,}197 \\
Node creator                        & 852 \\
Visualize / report / extractor      & 703 / 536--672 / 449 \\
\bottomrule
\end{tabular}
\end{minipage}%
\hfill
\begin{minipage}[t]{0.45\textwidth}
\centering
\begin{tabular}{lr}
\toprule
Injected schema & Tokens \\
\midrule
MuSiQue (FinReflectKG)      & 2{,}790 \\
HotpotQA (FinReflectKG)     & 2{,}039 \\
TriviaQA (FinReflectKG)     & 2{,}733 \\
OpenIE (each corpus)        & [..] \\ %
Per-corpus context prompt   & 444--574 \\
\bottomrule
\end{tabular}
\end{minipage}
\caption{Fixed per-call costs (exact, tokenized from the prompts).}
\label{tab:fixed-costs}
\end{table}

Shared fusion constants: RRF $k=60$, dense weight $=2.0$, sparse weight
$=1.0$. The adaptive metadata filter is \emph{None} for all three corpora
(chunks carry no metadata, so retrieval is unscoped). Qdrant dense-field
distance is cosine, with a separate collection per corpus. A
256-dimensional multivector setting is defined in configuration but
\emph{not applied} on these pipelines --- the dense field uses the
embedding model's native width.

\subsubsection{LLM Sampling Parameters}
Agent default: temperature $=0.1$, top-$p=0.95$, thinking enabled. Per-stage
call-site overrides:

Other shared pipeline defaults: triplet enrichment enabled, evidence
threading disabled, subplan expansion enabled, 3 explorer turns, and no
adaptive metadata filter. Subgraph scoping confines every
explorer/generator Cypher query to the chunks retrieved for that query,
via a shared identifier list bound at the top of each query.

\begin{table*}[!t]
\centering
\small
\begin{tabular}{llrrr}
\toprule
Corpus & Graph family & Context tok (mean/p90) & Answer tok (mean) & Total tok/rec \\
\midrule
MuSiQue  & FinReflectKG & 3{,}282 / 5{,}940   & 136 & 12{,}085 \\
MuSiQue  & OpenIE       & 5{,}753 / 15{,}826  & 131 & 9{,}723  \\
\midrule
HotpotQA & FinReflectKG & 3{,}637 / 7{,}618   & 117 & 9{,}924  \\
HotpotQA & OpenIE       & 2{,}449 / 5{,}815   & 143 & 6{,}287  \\
\midrule
TriviaQA & FinReflectKG & 32{,}632 / 78{,}394 & 103 & 38{,}815 \\
TriviaQA & OpenIE       & 4{,}873 / 12{,}650  & 125 & 8{,}949  \\
\bottomrule
\end{tabular}
\caption{Persisted-footprint comparison by graph-construction family
(same measurements as Table~\ref{tab:persisted-footprint}, regrouped by
family).}
\label{tab:footprint-graph-family}
\end{table*}

\section{Token-Cost Analysis}
\label{app:tokencost}
Three views follow: the persisted footprint of the stored result files,
the fixed per-call prompt costs, and a comparison of the footprint by graph
family. The footprints are measured from stored outputs, and the prompt
costs are tokenized from the prompts themselves.

\subsection{Persisted Footprint (Result JSONLs)}
\label{app:persisted-footprint}

Per-record token footprint of the stored artifacts --- a well-defined
proxy for answer-stage I/O and for the cost of any re-processing pass:

\begin{table}[!t]
\centering
\footnotesize
\setlength{\tabcolsep}{2.5pt}
\begin{tabular}{@{}llrrrr@{}}
\toprule
 & & \multicolumn{2}{c}{Context tok} & Answer & Total \\
\cmidrule(lr){3-4}
Corpus & Graph & mean & p90 & tok & tok/rec \\
\midrule
MuSiQue  & FinReflectKG & 3{,}282  & 5{,}940  & 136 & 12{,}085 \\
         & OpenIE       & 5{,}753  & 15{,}826 & 131 & 9{,}723 \\
\midrule
HotpotQA & FinReflectKG & 3{,}637  & 7{,}618  & 117 & 9{,}924 \\
         & OpenIE       & 2{,}449  & 5{,}815  & 143 & 6{,}287 \\
\midrule
TriviaQA & FinReflectKG & 32{,}632 & 78{,}394 & 103 & 38{,}815 \\
         & OpenIE       & 4{,}873  & 12{,}650 & 125 & 8{,}949 \\
\bottomrule
\end{tabular}
\caption{Persisted footprint per record, by corpus and graph family. Each
run has 1{,}000 records; the six runs total 85.78M tokens.}
\label{tab:persisted-footprint}
\end{table}

\textbf{Composition (estimated: field-level proportions from the full
runs, rescaled to the 85.78M-token total).} The retrieved
graph-row context fed to the answerer is the dominant footprint at
$\sim$60.0M tokens ($\sim$70\%). Generated Cyphers and plans account for
$\sim$17.2M ($\sim$20\%); explorer and planner working notes account for
$\sim$6.9M ($\sim$8\%). The actual generated answers total only
$\sim$0.6M tokens ($\sim$0.7\%) --- answer \emph{output} is negligible;
cost is overwhelmingly input.

\subsection{Fixed Per-Call Costs}
\label{app:per-stage-model}

Every agent call carries a fixed prompt template and the graph's
introspected schema, in addition to context that varies with the question.
Table~\ref{tab:fixed-costs} gives the fixed parts, tokenized from the
prompts; the variable context is not included.

\subsection{Footprint by Graph Family}
\label{app:footprint-graph-family}

This comparison regroups Table~\ref{tab:persisted-footprint} by graph
construction: schema-guided FinReflectKG versus open-vocabulary OpenIE,
queried by the same pipeline with the same per-corpus configuration. It
describes cost, not accuracy; accuracy by graph family is reported in the
main paper, and component ablations in
Appendix~\ref{app:component-ablations}.

\begin{table*}[!t]
\centering
\small
\setlength{\tabcolsep}{5pt}
\begin{tabular}{@{}llccc@{\hspace{14pt}}ccc@{}}
\toprule
 & & \multicolumn{3}{c}{FinReflectKG: passes} & \multicolumn{3}{c}{OpenIE: passes} \\
\cmidrule(lr){3-5}\cmidrule(l){6-8}
Configuration & Retries & 1 & 3 & 5 & 1 & 3 & 5 \\
\midrule
Generator only & 1 &  12.0 &  19.0 &  21.0 &  6.0 &  17.0 &  19.0 \\
               & 3 &  18.0 &  35.0 &  42.0 &  19.0 &  39.0 &  49.0 \\
               & 5 &  19.0 &  42.0 &  50.0 &  25.0 &  51.0 &  59.0 \\
\midrule
Explorer +      & 1 & 27.0 & 47.0 & 56.0 & \textbf{32.0} & \textbf{45.0} & \textbf{51.0} \\
\quad Generator & 3 & \textbf{34.0} & 65.0 & \textbf{76.0} & \textbf{41.0} & \textbf{66.0} & 68.0 \\
                & 5 & \textbf{53.0} & \textbf{81.0} & \textbf{88.0} & \textbf{42.0} & 66.0 & 71.0 \\
\midrule
Explorer +      & 1 & \textbf{30.0} & \textbf{58.0} & \textbf{64.0} & 21.0 & \textbf{45.0} & 49.0 \\
\quad Planner + & 3 & 29.0 & \textbf{69.0} & 74.0 & 36.0 & 63.0 & \textbf{73.0} \\
\quad Generator & 5 & 31.0 & 60.0 & 66.0 & 41.0 & \textbf{70.0} & \textbf{74.0} \\
\bottomrule
\end{tabular}
\caption{Component ablations on both graph families: EM (\%) by
configuration, Generator retries per query, and full pipeline passes, on the
same stratified sample. Generator only at 1 retry and 1 pass is the
Direct-Cypher agent; at larger budgets it is ArcRAG's Generator without
exploration or planning. For each graph family and retry and pass budget,
the best configuration is in bold and the lowest in red; at 1 retry and
3 passes on OpenIE, two configurations tie for best.}
\label{tab:component-ablations}
\end{table*}

\begin{figure*}[t]
    \centering
    \includegraphics[width=\textwidth,height=0.45\textheight,keepaspectratio]{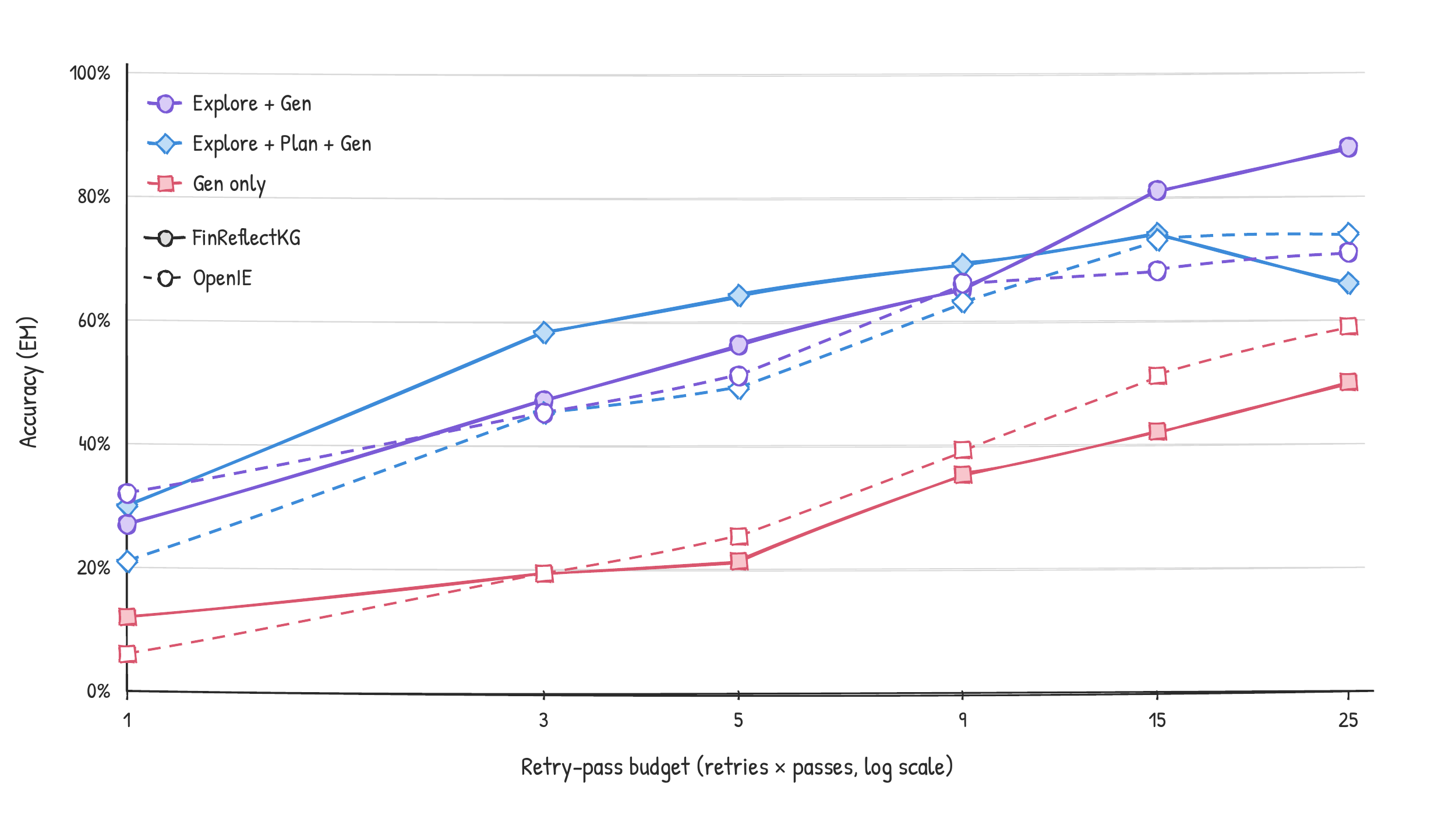}
    \caption{Accuracy (\%) of the three configurations as the retry--pass
    budget (retries $\times$ passes, log scale) grows. Budgets 3, 5, and 15
    can each be reached with two settings (e.g.\ 1 retry $\times$ 5 passes
    or 5 retries $\times$ 1 pass); Table~\ref{tab:component-ablations}
    reports every setting separately.}
    \label{fig:component_ablations}
\end{figure*}

\paragraph{Which family retrieves more context?} The direction is not
consistent across corpora:
\begin{itemize}
\item \textbf{MuSiQue} --- OpenIE retrieves \emph{more} context than
FinReflectKG: 5{,}753 vs.\ 3{,}282 mean tokens ($\sim 1.75\times$).
One possible explanation is that, without typed anchors to narrow the
search, retrieval over the open-vocabulary graph returns more candidate
rows; we have not tested this.
\item \textbf{HotpotQA} --- the direction reverses: FinReflectKG retrieves
\emph{more} context than OpenIE: 3{,}637 vs.\ 2{,}449 mean tokens
($\sim 1.49\times$).
\item \textbf{TriviaQA} --- FinReflectKG is the outlier by a wide margin:
32{,}632 vs.\ 4{,}873 mean tokens ($\sim 6.7\times$), with an even wider
p90 gap (78{,}394 vs.\ 12{,}650). The mean/p90 ratio within FinReflectKG
itself ($\sim 2.4\times$) is consistent with a heavy right tail, in which a subset of queries
run toward the iteration cap and accumulate large context.
\end{itemize}

\paragraph{Context is not the same as total footprint.} On
\textbf{Total tok/rec}, FinReflectKG is consistently the more expensive
family across \emph{all three} corpora --- 12{,}085 vs.\ 9{,}723
(MuSiQue), 9{,}924 vs.\ 6{,}287 (HotpotQA), 38{,}815 vs.\ 8{,}949
(TriviaQA) --- even for MuSiQue, where FinReflectKG's own retrieved
context is the \emph{smaller} of the two. That means non-context fields
(subplans, exploration reports) make up a larger share of the
FinReflectKG footprint than the OpenIE footprint for MuSiQue specifically.

\section{Component Ablations}
\label{app:component-ablations}

Table~\ref{tab:component-ablations} and
Figure~\ref{fig:component_ablations} vary which stages run and how much
budget they receive, on both graph families. All runs use the same
stratified sample of 100 questions: 33 from MuSiQue, 34 from HotpotQA, and
33 from TriviaQA. Two budgets are varied: \emph{retries}, the number of
attempts the Generator is allowed per query, and \emph{passes}, the number
of full iterations of the retrieve--answer loop. The figure plots accuracy
against their product to show the overall trend; the table separates the
two. We read the results as a search problem. The Explorer expands the
frontier, establishing which relations and paths actually exist around the
anchors; the Planner acts as a heuristic, ranking candidate paths and
committing to some of them; each Generator retry tests a candidate by trial
and error; and each pass restarts the search from a follow-up question.

\paragraph{Direct-Cypher baseline and Generator alone.} With one retry and
one pass, the Generator alone is a Direct-Cypher agent: it compiles Cypher
from the question and the introspected schema in a single attempt, with no
exploration or plan, and answers 12.0\% of questions on FinReflectKG and
6.0\% on OpenIE. With more retries and passes it is no longer a plain
text-to-Cypher baseline but ArcRAG's Generator without exploration or
planning: its refinement loop loosens or tightens a query in response to
the result, and further passes restart retrieval from a follow-up question.
This raises accuracy to 50.0\% and 59.0\% at 5 retries and 5 passes. That it
does better on OpenIE at high budgets suggests that, with a single
relationship type, almost any query pattern is valid, so refinement becomes
a search over predicate strings; we have not tested this explanation.

\paragraph{Exploration.} Adding the Explorer improves on the Generator alone
at every budget on both graph families: by 15--39 points on FinReflectKG and
by 12--32 on OpenIE. The relative gain is largest where the schema says
least: at one retry and one pass on OpenIE, exploration raises accuracy from
6.0\% to 32.0\%. As budgets grow on OpenIE, the Direct-Cypher agent closes
part of the gap, and exploration's advantage narrows to 12 points at
5 retries and 5 passes.

\paragraph{Planning.} The Planner's contribution depends on the graph. On
FinReflectKG it improves on Explorer + Generator when the Generator has a
single retry (by 3--11 points) and gives the best accuracy at every
retry--pass budget up to 9; at 5 retries Explorer + Generator is ahead by
21--22 points, which suggests that committing to a plan narrows the search
the Generator would otherwise perform. On OpenIE the pattern reverses: the
Planner is no better than Explorer + Generator at budgets up to 9, and 11
points behind at one retry and one pass, while at larger budgets it is ahead
by 3--5 points, within sampling noise at this sample size. One possible
reason is that on an open-vocabulary graph a plan commits to specific
predicate strings, and a single retry cannot repair a wrong commitment; we
have not tested this explanation either.

\begin{figure*}[!t]
    \centering
    \includegraphics[width=\textwidth,height=0.45\textheight,keepaspectratio]{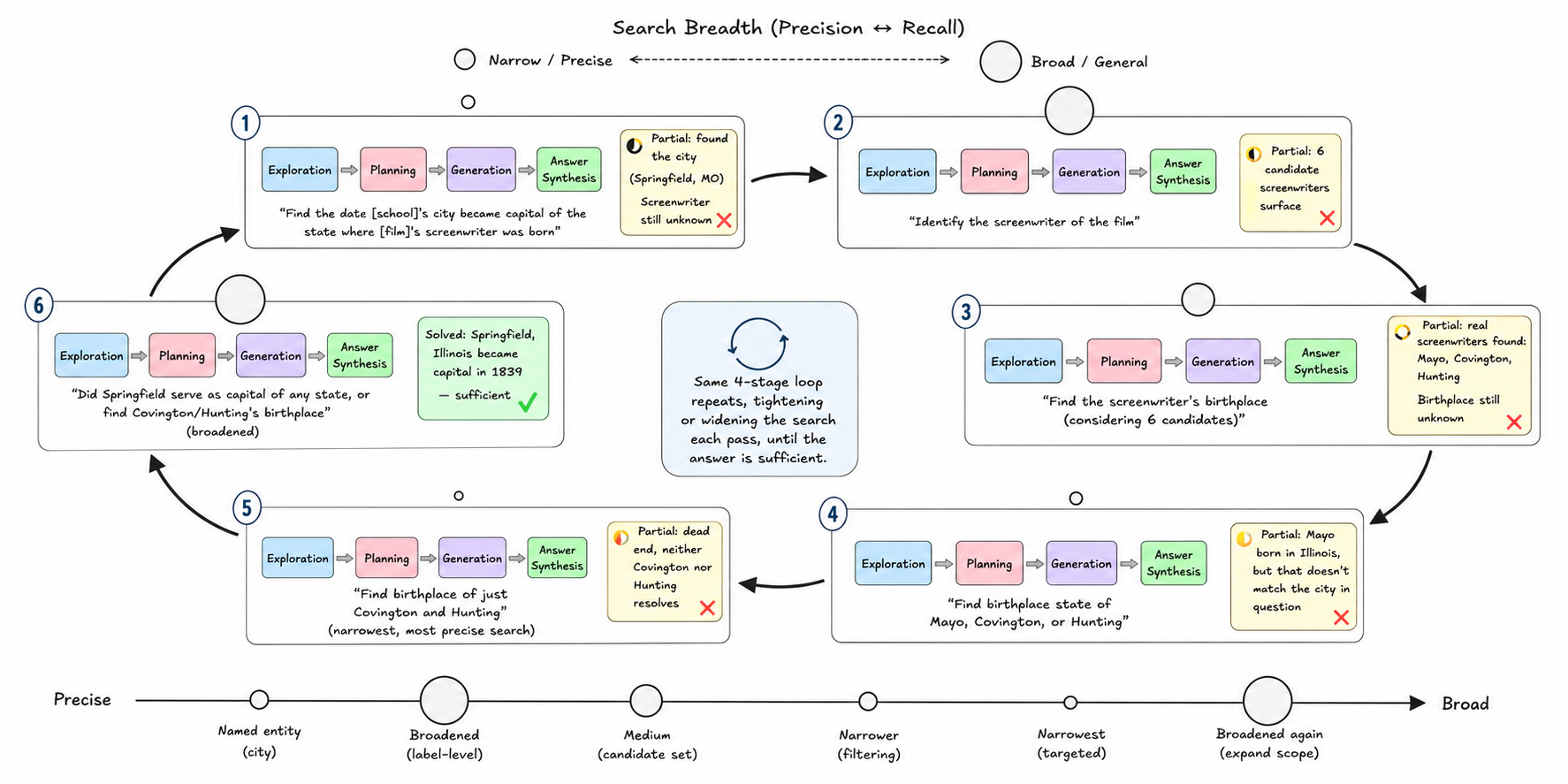}
    \caption{Iterative search process with adaptive search breadth.}
    \label{fig:iterative_search_loop}
\end{figure*}

\paragraph{Summary.} Query-local exploration is the component that
transfers: it improves on direct Cypher generation at every budget on both
constructions. The Planner is a graph-dependent heuristic. On a typed graph
it tells the Generator which explored path to try, which matters most when
retries are scarce; on an open-vocabulary graph its guidance adds little at
small budgets and a few points at large ones.

\section{Qualitative Examples and Error Analysis}
\label{app:qualitative}

\subsection{Example: Multi-Hop Self-Correction (MuSiQue, 4-hop)}
\label{app:qual-musique}

\paragraph{Question.} \textit{Find the date when the city that is the
location of Greenwood Laboratory School became the capital of the state
in which the screenwriter of the film ``The Poor Boob'' was born.}

\paragraph{Reasoning trace.}
Figure~\ref{fig:iterative_search_loop} visualizes the six passes. In \textbf{Pass 1}, the system resolves
\textit{Greenwood Laboratory School} to \textit{Springfield, Missouri},
successfully identifying the city but leaving the film's screenwriter
unknown. Since only part of the reasoning chain is resolved, answer
synthesis marks the pass as insufficient.

In \textbf{Pass 2}, the planner broadens the search from a precise
entity lookup to a wider search over film credits, retrieving six
candidate screenwriters. Although additional evidence is obtained, none
can yet be confirmed as the correct writer.

In \textbf{Pass 3}, the search narrows again by investigating the
birthplaces of the candidate set. This eliminates several false leads
and identifies the actual credited writers---Margaret Mayo,
Z.~Wall Covington, and Gardner Hunting---but their birthplaces remain
unresolved.

In \textbf{Pass 4}, the planner further narrows the search to the
birth states of these three writers. Margaret Mayo is found to have
been born in Illinois, while the remaining candidates provide no useful
evidence. This partially resolves the reasoning chain but creates a
conflict with the earlier Missouri anchor.

In \textbf{Pass 5}, the system performs its most targeted search,
attempting to resolve only Covington's and Hunting's birthplaces.
Neither search succeeds, producing no additional evidence.

Finally, in \textbf{Pass 6}, the planner broadens the search once
more, abandoning the exhausted candidate-specific searches and instead
querying whether Springfield served as the capital of any state.
Combining this with the Illinois evidence from Margaret Mayo, the system
retrieves that \textit{Springfield, Illinois became the state capital in
1839}, completing the reasoning chain and producing a sufficient answer.

\paragraph{Final answer.}
\textit{1839} (correct).

\paragraph{Discussion.}
This example demonstrates the iterative search strategy illustrated in
Figure~\ref{fig:iterative_search_loop}. Rather than repeatedly attempting
the same retrievals, each pass contributes partial progress by
either discovering new entities, eliminating incorrect hypotheses, or
adapting the search scope from precise entity-level retrieval to broader
graph exploration. Evidence accumulates across passes and incrementally
refines the active reasoning chain until the sixth pass has sufficient
evidence to answer the question.

\section{Compute and Reproducibility}
\label{app:repro}
All language models and the embedder were accessed through an
OpenAI-compatible inference API; no models were trained or fine-tuned, and
no local GPUs were used. The pipeline is implemented in Python using the
OpenAI client library for model calls, the Neo4j Python driver for Cypher
queries over Memgraph via the Bolt protocol, and the Qdrant client for
hybrid dense and BM25 retrieval ([versions]). The six main runs (three
datasets, two graph families, 1{,}000 questions each) persist 85.78M tokens
(Appendix~\ref{app:tokencost}), and the component ablations comprise 54 runs
over the same 100-question sample. All settings are listed in
Appendix~\ref{app:hyperparams}; agents run near-greedy (temperature
0.0--0.2), the answer model runs at temperature 1.0, and each configuration
was run once.

\end{document}